\documentclass[letterpaper]{article}
\usepackage{aaai}
\usepackage{times}
\usepackage{helvet}
\usepackage{courier}
\usepackage{amsmath}
\usepackage{amssymb}
\usepackage{mathtools}
\usepackage{amsthm}
\usepackage{booktabs}
\usepackage{multirow}
\usepackage{tabularx}
\usepackage{cite}
\usepackage{colortbl}  
\usepackage{natbib}
\usepackage{xcolor}
\definecolor{mydarkblue}{rgb}{0.0, 0.2, 0.6}
\definecolor{mylightgreen}{rgb}{0.6, 0.9, 0.6}
\definecolor{mylightgray}{rgb}{0.9, 0.9, 0.9}

\begin{document}
%
\title{Beyond IID: Optimizing Instruction Learning from the Perspective of Instruction Interaction and Dependency}
\author{Hanyu Zhao$^*$, Li Du\thanks{Equal contribution.} \thanks{Corresponding Author}, Yiming Ju, Chengwei Wu, Tengfei Pan \\
Beijing Academy of Artificial Intelligence, Beijing, China \\
\{hyzhao, duli, ymju, cwwu, tfpan\}@baai.ac.cn\\
}

\maketitle
\begin{abstract}

With the availability of various instruction datasets, a pivotal challenge is how to effectively select and integrate these instructions to fine-tune large language models (LLMs). Previous research mainly focuses on selecting individual high-quality instructions. However, these works overlooked the joint interactions and dependencies between different categories of instructions, leading to suboptimal selection strategies.
Moreover, the nature of these interaction patterns remains largely unexplored, let alone optimize the instruction set with regard to them.
To fill these gaps, in this paper, we: (1) systemically investigate interaction and dependency patterns between different categories of instructions, (2) manage to optimize the instruction set concerning the interaction patterns using a linear programming-based method, and optimize the learning schema of SFT using an instruction dependency taxonomy guided curriculum learning. 
Experimental results across different LLMs demonstrate improved performance over strong baselines on widely adopted benchmarks.

\end{abstract}


\section{Introduction}

    
Supervised fine-tuning (SFT) is the key to aligning large language models (LLMs) with human beings, enabling them to complete various downstream tasks and adapt to specific domains such as healthcare and finance \citep{zhao2023survey}. The effectiveness of the SFT process relies on a high-quality instruction set, so as to ensure the performance of LLMs \citep{xu2023wizardlm,wang2023self,longpre2023flan}. 
Temporarily, with the availability of various instruction sets, a new challenge has been raised, i.e., how to select and integrate existing datasets to obtain an optimized instruction set.
To address this issue, previous research typically works by selecting and combining individual ``high-quality'' instructions. 
Then often construct proxy indicators to evaluate different aspects of quality, such as factual correctness, complexity, and informativeness. Then the raw instruction set could be refined by selecting instructions with the highest relative quality scores \citep{latif2024fine,zhao2023preliminary,lu2023instag,li2024quantity}.

However, emerging evidence \citep{dongabilities, yuan2023hype} and our analyses indicate that complex correlation and dependency relationships exist between different categories of instructions. Therefore, considering the quality of individual instructions alone can be a suboptimal approach for building a fine-tuning instruction set.
Research indicates that these categories are interrelated; incorporating one category of instructions may \emph{enhance or diminish} the model's performance in others \citep{dong2023abilities,huang2023towards}. Additionally, the skills required for different tasks often form hierarchical taxonomies. For instance, solving a bioinformatics problem requires both biological knowledge and coding skills.
Consequently, instructions are interconnected and collectively influence model performance. Ignoring these correlations can reduce the efficiency of instruction selection, as incorporating one category of instruction may even degrade the model's performance in another category.
Moreover, the dependency between skills necessitates that models acquire foundational knowledge before progressing to more complex tasks; otherwise, the effectiveness of instruction tuning will be compromised \citep{longpre2023flan}.


Hence, it is crucial to account for these joint effects for optimizing the instruction set. 
However, two main challenges remain unaddressed: (1) The potential correlation and dependency patterns are largely unknown; (2) How to optimize the instruction set while considering these correlation and dependency patterns remains an unexplored area. 
\begin{figure*}[t]
\centering
\includegraphics[width=0.9\textwidth]{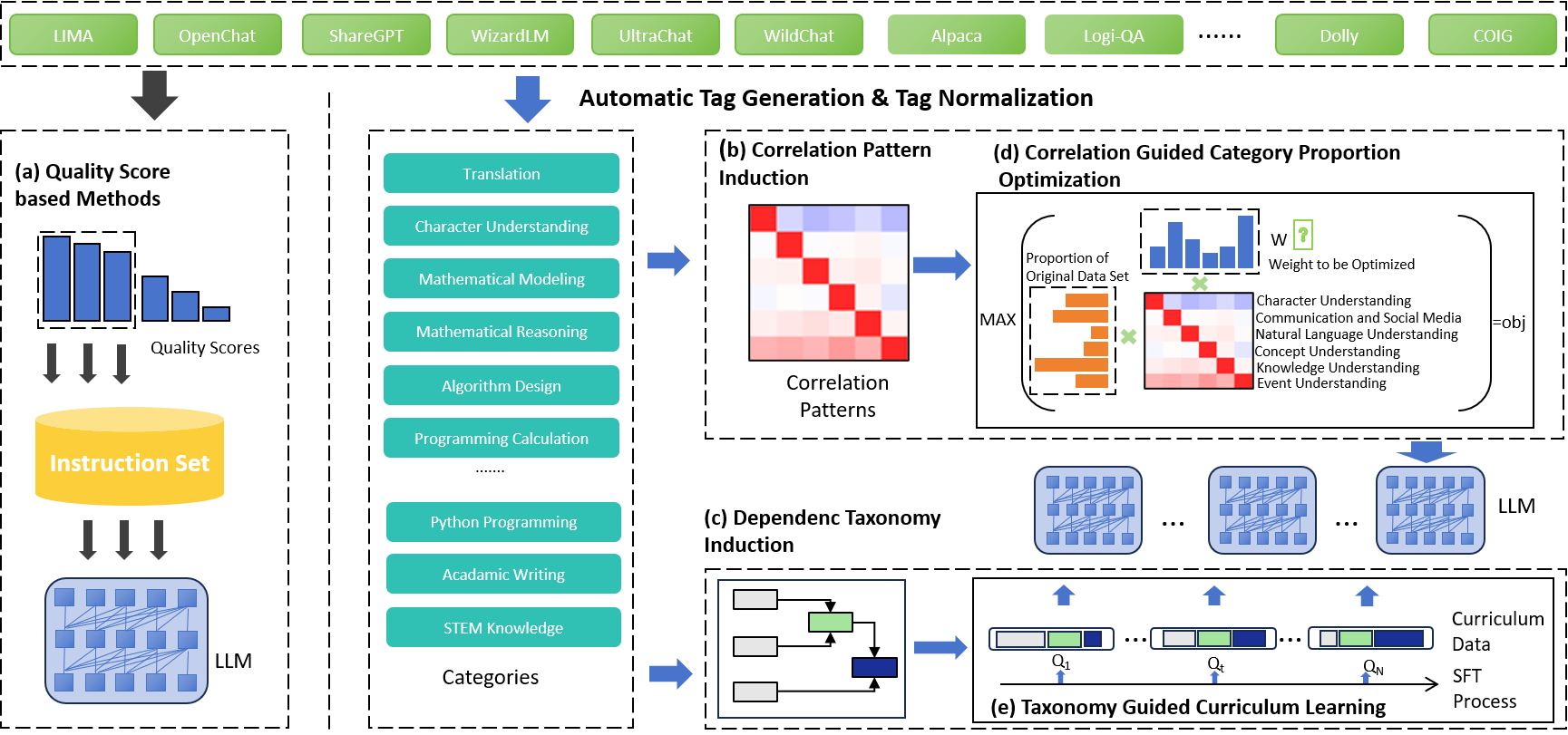}
\setlength{\abovecaptionskip}{-0.1cm}
\caption{Framework of our work. Baseline methods selection instructions using quality scores (a). In this paper, we first induce the correlation pattern (b) and dependency taxonomy (d), then optimize the instruction set collection concerning the correlation (c) and dependency taxonomy (e).}
\label{fig:main_frame}
\end{figure*}
To fill these gaps, as Figure~\ref{fig:main_frame}~(b)-(e) shows, we systemically investigated the correlation patterns between different categories of instructions, and induced an ability taxonomy of instructions based on causal interventions on the distribution of the instruction set. 
Then, with the guidance of the correlation patterns and dependency taxonomy, we optimized the instruction set by adjusting the proportion of different categories of instructions, and arranging the order of learning different categories of instructions.

Specifically, we construct an automatic tagging system to assign the instruction with tags describing the detailed capability and knowledge required to complete this instruction. 
With the tags, we do interventions in the dataset distribution by adding or removing instructions with certain tags, so that we can observe how the LLM's performance changes with the incorporation of each category of instructions, as well as how the performance of the LLM on one category of instruction changes depending on another category of instructions.  
Given the correlation patterns, we managed to optimize the proportion of different categories of instructions by turning it into an effect-equivalence-based linear programming problem. Furthermore, we propose a dependency taxonomy-based curriculum learning \citep{wang2021survey} method to rearrange the learning order of categories of instructions. We release the code and dataset at  https://github.com/BAAI-DIPL/sft-set-optimization-via-instruction-interaction-and-dependency.

Our experiments demonstrate extensive correlations and dependencies among various categories of instruction data, particularly between reasoning-related and commonsense memorization tasks. Mathematics and coding also emerge as foundational elements for LLMs in executing general domain tasks. By leveraging these correlations and dependencies, we applied guided optimization and curriculum learning methods, resulting in improved performance across different LLMs, including Qwen \citep{yang2024qwen2} and Llama \citep{dubey2024llama}, compared to state-of-the-art baselines on widely recognized benchmarks. Such results in turn support the reasonability of our analysis method and induced instruction interaction patterns.


\section{Causal Intervention based Instruction Correlation Analysis and Ability Taxonomy Induction}
In the SFT stage, the LLM is trained to learn given instruction set $\mathcal{D}=\{\mathcal{C}_i\}_{i=1}^n$, where $\mathcal{C}_i$ is the $i$th category of instructions. With the maximum likelihood estimation and independent identical distribution (iid) assumption, the training objective could be formalized as $\max_{\theta} \prod_i [P(\mathcal{C}_{ij})]$, where $\mathcal{C}_{ij}$ is the $j$th instruction of $\mathcal{C}_i$. 

Due to the inherent correlation and dependency between the knowledge and skills involved in different categories of instructions, the distribution of instruction set $\mathcal{D}$ should be characterized using an joint distribution $P(\mathcal{C}_1, \dots, \mathcal{C}_i, \mathcal{C}_{N})$, and
$P(\mathcal{C}_1, \dots, \mathcal{C}_i, \dots, \mathcal{C}_{N}) \neq \prod_i P(\mathcal{C}_{ij})$. Such discrepancy would lead to ineffectiveness and inefficiency of the SFT process, as: (1) Under the iid assumption, the training objective deviates from the actual distribution of instruction set; (2) The joint distribution could be decomposed into a sequential manner as $P(\mathcal{C}_1, \dots, \mathcal{C}_i, \dots, \mathcal{C}_{N})=\prod P(\mathcal{C}_k|\mathcal{C}_j, \dots)$, which indicates a sequential learning schema, i.e., advanced skills could be learned only enough preliminary knowledge is equipped. Nevertheless, in the current SFT schema, different categories of instructions are randomly distributed in the whole epoch, making the advanced skills been trained before equipped with enough prior knowledge. This limits the efficiency of the SFT process. Thus, it would be necessary to consider the optimization of instruction set \textbf{beyond the iid assumption}.

However, the category of instructions are unknown priorly. Thus, to induce the correlation pattern and dependency taxonomy between different categories of instructions, we first systematically collect publicly available high-quality instructions and design an ability tagging system to automatically confer instruction a set of tags, which describes the ability and knowledge necessary for completing the instruction. So that we can categorize the instructions with the tags. Then based on the category tags, by adding or removing certain categories of instructions (Cause), we could obtain how the performances of other categories of instructions change (Effect) brought by such intervention, and induce the correlation and dependency pattern. 

\subsection{Instruction Collection and Automatic Ability Tagging System}

The prerequisites of analyzing the relationship patterns between different categories of instructions are collecting a large enough instruction set and elucidating the category distribution of the instruction set. To this end, we comprehensively collect currently available high-quality open-source instruction sets, and build an automatic tagging system, to confer each instruction tag(s) about the main skill or knowledge necessary for completing the instruction. For example, as shown in Table~\ref{tab:tagging_dat_example}, the LLM should have both the programming ability and STEM knowledge (biology) to fulfill the requirement described by the instruction.

\begin{table}
    \centering
    \small
    \begin{tabular}{p{5.5cm}|p{2cm}}
    \toprule
    Instruction  & Category Tag  \\
    \midrule
    Write a piece of code to count the number of different types of nucleotide bases in a DNA sequence:  &  [STEM Knowledge', 'Program Ability']\\
    \bottomrule
    \end{tabular}
    \caption{Examples of the tags of instruction.}
    \label{tab:tagging_dat_example}
\end{table}




We systematically collected the available high-quality instruction sets constructed by manual annotation, GPT4 \cite{achiam2023gpt} or GPT 3.5. After quality filtering and de-duplication, we collect a large-scale instruction collection with ~9 million instructions. 


Given the vast scale instructions set, we construct an LLM-based tagging system to automatically confer each instruction a set of tags. Specifically, we employ Qwen-1.5-72B-Instruction \citep{yang2024qwen2} as the tagger, and guide it to generate tags through prompts. 
Whereas the LLM may describe the same ability or knowledge using different expressions, making the normalization of tags necessary. 
To address this issue, we combine the tags with high semantic similarity. Specifically, we first obtain the semantic representation of the tags using a text embedding model BGE \citep{xiao2023c}. Then semantically similar tags are recognized if their cosine similarity of embeddings is larger than an empirical threshold $ \lambda=0.85 $. For a set of semantically similar tags, they are normalized to the one with the highest frequency among them \citep{hahsler2019dbscan}. After normalization, about 21,000 tags are left. More details about the instruction collection and cleaning, tag construction, normalization, and results of tagging are described in the Appendix. 

It would be impractical to investigate the correlation and dependency of all the 21,000 categories of instructions. Thus, as listed in Table~\ref{tab:cate_list}, we manually choose 29 categories of instructions according to frequency and importance, which cover the main tasks and abilities across the Math, Coding, QA, Commonsense Reasoning, Natural Language Processing and Understanding, together with Dialogue and Applications.

\begin{table}
    \centering
    \small
    \begin{tabular}{p{1.7cm}|p{5.5cm}}
    \toprule
    \textbf{Domain}  & \textbf{Category List}\\
    \midrule
    Math & ['Math Reasoning', 'Mathematical Modelling', 'Arithmetic Calculation', 'Data Process and Analysis']\\
    \midrule
    Coding & ['Python', 'Java', 'Programm Ability', 'Coding Algorithm'] \\
    \midrule
    QA & ['STEM Knowledge QA', 'Humanities \& Social Sciences QA',  'Commonsense Understanding', 'Open Domain QA']\\
    \midrule
    Commonsense Reasoning & ['Commonsense Reasoning', 'Concept Understanding', 'Logical Reasoning'] \\
    \midrule
    NLP \& NLU & ['Information Extraction', 'Sentiment Analysis', 'Story Understanding', 'Text Classification', 'NLU', 'Textual Summarization', 'Translation', 'Event Understanding']\\
    \midrule
    Dialogue \& Applications & ['Multiturn Dialogue', 'Communication \& Social Media', 'Character Understanding and Role-Playing', 'String Process', 'Academic Writing', 'Creative Writing'] \\
    \bottomrule
    \end{tabular}
    \caption{Categories of instructions included for analysis.}
    \label{tab:cate_list}
\end{table}

\subsection{Causal Intervention based Instruction Correlation Analysis}



Previous research suggests that instructions from different domains and tasks are interconnected. After incorporating one category of instructions, after SFT, performance across other categories would also be influenced. The source of such correlations could be rather complex \cite{dong2023abilities,huang2023towards}.
In this paper, rather than investigating the source of such correlation, we focus on systematically inducing the patterns of correlation, so as to directly guide the optimization of instruction sets. 

There are various potential methods for quantifying such correlation. In this paper, we propose a \emph{effect equivalence coefficient} to quantify the correlation between the $i$th and $j$th category of instruction:
\begin{equation}
    \gamma_{ij}^{M}= \text{Avg} \left( \frac{\rho[M^{\cup \textcolor{red}{\tilde{\mathcal{C}}_i}}(C_{\text{eval},jk})]-\rho[M(C_{\text{eval},jk})]}{\rho[M^{\cup \textcolor{blue}{\tilde{\mathcal{C}}_j}}(C_{\text{eval},jk})]-\rho[M(C_{\text{eval},jk})]} \right)
\end{equation}
where $M$ is a base model, $\mathcal{D}$ is an already existing instruction set, $M^{\cup \textcolor{red}{\tilde{\mathcal{C}}_i}}$ and $M^{\cup \textcolor{blue}{\tilde{\mathcal{C}}_j}}$ is obtained by finetuning $M$ on $\mathcal{D} \cup \textcolor{red}{\tilde{\mathcal{C}}_i}$ and $\mathcal{D} \cup \textcolor{blue}{\tilde{\mathcal{C}}_j}$, respectively, $\tilde{\mathcal{C}}_i$ are instructions of category $i$ out of $\mathcal{D}$. $M(C_{\text{eval},jk})$ is the output of $M$ on the $k$th instruction of category $j$ on the evaluation set $(C_{\text{eval},jk})$, $M^{\cup \textcolor{red}{\tilde{\mathcal{C}}_i}}(C_{\text{eval},jk})$ similarly.  $\rho(\cdot)$ is a performance evaluation function.
Thus, heuristically, \textbf{the effect equivalence coefficient $\gamma_{ij}^{M}$ measures one instruction of category $i$ ``equals'' how many instructions of category $j$ on average, with the existence of an existing instruction set $\mathcal{D}$}. Thus, a larger $\gamma_{ij}^{M}$ indicates a stronger correlation.

The reasons for measuring the effect equivalence coefficient with regard to $\mathcal{D}$ are twofold: (1) in practical scenarios, it often involves incorporating a certain category of instructions into the existing instruction set to enhance the LLM's capabilities in this area. Thus, investigating the correlation pattern under such a scenario would guide evaluating the influence of category proportion adjustments. (2) If inducing $\rho_{ij}$ by removing $\mathcal{C}_i$ from $\mathcal{D}$, it would be hard to elucidate whether the performance change is due to the absence of necessary preliminary knowledge in $\mathcal{C}_i$, or the correlations between the instructions.  
The base instruction set $\mathcal{D}$ at each time with different categories of instructions evenly distributed. The list of instruction data categories is shown in Table 1. We set the performance evaluation function $\rho(\cdot)$ as the log-likelihood of the response corresponding to $C_{\text{eval},jk}$. 

One remaining issue is that $\gamma_{ij}^{M}$ depends on the choice of $M$. Actually, due to the similarity in the capabilities of frequently used open-source LLMs, the effect equivalence coefficients induced by different models could be rather similar. In this section, we demonstrate the analysis results obtained using $M=$Qwen-1.5-7B. More results based on other LLMs such as LLama-3-8B \citep{dubey2024llama} are provided in the Appendix. Moreover, we argue that due to the similarity in the pretraining corpus, the results based on 7B-sized models could be scaled up to models with larger sizes. 



\subsubsection{Experimental Settings}

We use llama3-8B and Qwen-1.5-7B as base models for inducing the correlations. 
Within the base instruction set and the evaluation set contain 1,000 and 500 instructions of each category, respectively. To induce the correlation patterns, at each time, 2,000 additional instructions belonging to one of the 29 categories are incorporated into the base instruction set. 

\begin{figure}[t]
\centering
\includegraphics[width=0.45\textwidth]{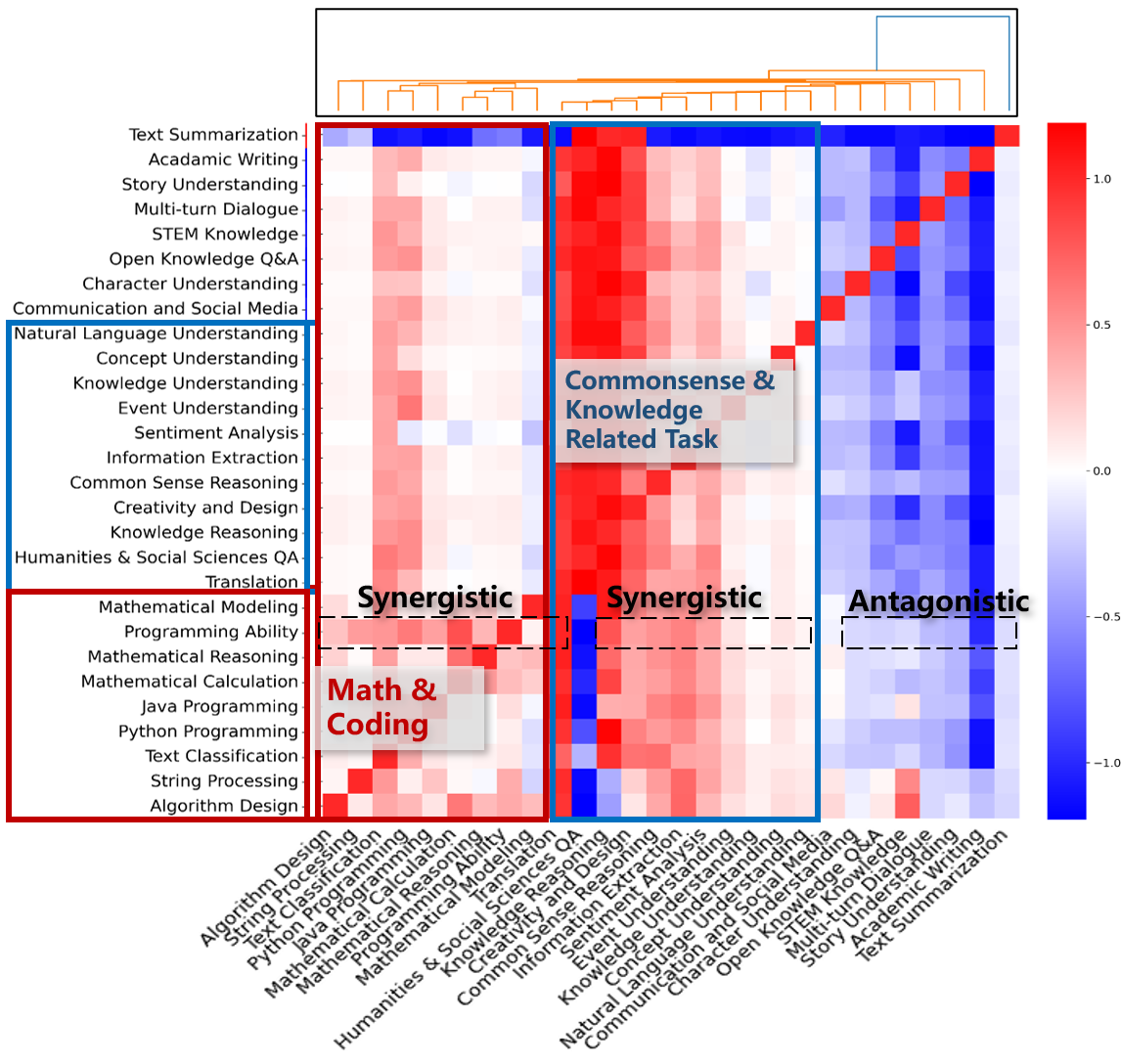}
\caption{The effect equivalence coefficients between different categories of instructions.}
\label{fig:equv_mat}
\end{figure}

\subsubsection{Analysis Results}

Figure~\ref{fig:equv_mat} and Figure 5 of the Appendix show the effect equivalence coefficients between different categories of instructions derived from Qwen and Llama, respectively. The $ij$th element of the matrix corresponds to $\gamma_{ij}^{M}$, i.e.,  one instruction of category $i$ ``equals'' how many instructions of category $j$ on average.  We observe that: (1) The existence of the correlation pattern is widespread, as it could be found across multiple categories, and upon different LLMs. Moreover, the correlation patterns show similarities across different LLMs. (2) Besides the positive relationships, the negative elements also account for a large proportion of the effect equivalence coefficients (e.g., The influence of Programming Ability on Text Summarization to Story Understanding). These suggest the wide existence of antagonistic relationships between different categories of instruction, i.e., incorporating one category of instruction would lead to performance degradation of another one. This highlights the necessity of optimizing the category proportion of the whole instruction set beyond filtering individual high-quality instructions or simply enlarging the scale of the instruction set. On the one hand, the synergistic categories may lead to redundancy of the instruction set, since the instructions could substitute each other to some extent; on the other hand, due to the existence of performance antagonistic effect, adding instructions of one category would jointly impact performance on the other categories, and vice versa. Especially when constructing domain models, often involves incorporating a large number of domain-specific or task-specific instructions. This further necessitates the careful arrangement of the types of instructions to ensure performance on other domains is not severely impacted.
Moreover, one category of instructions may have synergistic and antagonistic effects with other categories of instructions at the same time. 
For example, incorporating the Program Ability category could promote the performance of Math and Code related categories, meanwhile impacting the performance on STEM Knowledge or open Domain QA. 
Such complexity further increases the difficulty of category proportion optimization.
(2) According to the correlation patterns, using hierarchical clustering, the instruction categories could be further classified into two ``meta-groups'': I. A Symbolic Reasoning related group, including math and coding-related categories. II. A Commonsense Memorization-related group, including knowledge understanding, knowledge QA, etc. 
Heuristically, the categories within the meta-groups do share inherent similarities in the knowledge and ability required for completing these tasks, suggesting the effect equivalence coefficients can reflect the inner correlations between different instruction categories.    
Since both meta-groups are crucial for LLMs, it is necessary to optimize the category distribution of instruction sets to enhance both types of meta-capabilities simultaneously, concerning the performance antagonistic effects.

\subsection{Causal Intervention based Large Language Model Ability Taxonomy Induction}

Heuristically, human beings could not learn advanced knowledge before mastering the necessary preliminary knowledge. For example, it would be rather hard for a student to learn advanced math before he has acquired basic arithmetic calculations. Such a phenomenon inspires us to investigate whether the dependency also exists when LLMs learn different categories of instructions in the SFT process.  

To induce the dependency taxonomy of different instruction categories, given an instruction set $\mathcal{D}_{\text{train}}$, we sequentially remove one category of instructions $\mathcal{C}_i$ and obtain a set of ablation instruction sets $\{\mathcal{D}_{\text{train}}^{\backslash \mathcal{C}_i} \}_{i=1}^{N}$, $N$ is the total number of instructions, then compare the performance change of LLM fine-tuned on $\mathcal{D}_{\text{train}}$ with that fine-tuned on $\mathcal{D}_{\text{train}}^{\backslash \mathcal{C}_i}$, and that fine-tuned on $\mathcal{D}_{\text{train}}^{\backslash \mathcal{C}_j}$. This is because: (1) If the exclusion of $\mathcal{C}_i$ \emph{causes} significant performance degradation on another category of instructions $C_{j}$ (\emph{effect}), then it could be assumed that the LLM fails to learn $\mathcal{C}_j$ if $\mathcal{C}_i$ not exists; (2) If without $\mathcal{C}_j$, $\mathcal{C}_i$ could also be well learned, then the performance degradation of $\mathcal{D}_{\text{train}}^{\backslash \mathcal{C}_j} $ is not due to the synergistic effect between $\mathcal{C}_i$ and $\mathcal{C}_j$. Thus, $\mathcal{C}_j$ depends on $\mathcal{C}_i$.
Note that, compared to the observational-based dependency induction methods, the causal intervention-based method could provide the strongest evidence.

One remaining issue is how to define the ``significant'' performance degradation. To this end, we measure the performance of LLM on a certain category of instructions using the average PPL on the evaluation set, and employ a non-parametrical statistical test to measure the significance. Specifically, given LLMs $M$ and $M^{\backslash \mathcal{C}_i}$ fine-tuned on $\mathcal{D}_{\text{train}}$ and $\mathcal{D}_{\text{train}}^{\backslash \mathcal{C}_i}$ respectively, on the evaluation set, we can obtain the PPL on the $k$th instance of the $j$th category $\text{PPL}({C_{\text{eval},jk}})$ and $\text{PPL}^{\backslash \mathcal{C}_i}({C_{\text{eval},jk}})$. To compare whether $M^{\backslash \mathcal{C}_i}$ has a larger PPL than $M$ on $\mathcal{C}_j$, considering the complexity of the distribution of PPL, we test whether $\{\text{PPL}^{\backslash \mathcal{C}_i}({ C_{\text{eval},jk}})-\text{PPL}({C_{\text{eval},jk}})\}_{k=0}^{|C_{\text{eval},j}|}>0$ using the non-parametrical Wilcoxon signed-rank test. Furthermore, since given $N$ categories there would be $(N-1)^2$ times of statistical tests, the risk of the False Positive would be increased. Thus we further adjust the P-values using the Benjamini-Hochberg procedure and only keep the dependency relationships with an adjusted P-value smaller than 0.05.

\subsubsection{Experimental Settings}

Experiments are conducted on Llama3-8B and Qwen-1.5-7B, with the same base instruction set and category collection of instructions as the instruction correlation analysis. To induce the dependency taxonomy of different categories of instructions, one category of instructions is excluded from the base instruction set at each time. Note that since the number of instructions of each type is the same, the difference in the change of PPL after excluding different categories of instructions is not brought about by the difference in the number of instructions.

\begin{table}
    \centering
    \tiny
        \begin{tabular}{p{1.5cm}|p{6cm}}
        \toprule
        Subsequantial categories & \cellcolor{mydarkblue} \textcolor{white}{['Humanities \& Social Sciences QA', 'Commonsense Understanding', 'Open Domain QA', 'Communication \& Social Media',  'Character Understanding and Role-Playing', 'Creative Writing']} \\
        \midrule
        Intermediary Categories & \cellcolor{mylightgreen} ['Data Process and Analysis', 'STEM Knowledge QA', 'Commonsense Reasoning', 'Concept Understanding', 'Logical Reasoning', 'Information Extraction', 'Sentiment Analysis', 'Story Understanding', 'Text Classification', 'NLU', 'Text Summarization', 'Translation', 'Event Understanding', 'Multiturn Dialogue', 'String Process', 'Academic Writing'] \\
        \midrule
        Preliminary Categories & \cellcolor{mylightgray} ['Math Reasoning', 'Mathematical Modelling', 'Arithmetic Calculation', 'Python', 'Java', 'Programm Ability', 'Coding Algorithm'] \\
        \bottomrule
        \end{tabular}
    \caption{Dependency taxonomy between instruction categories.}
    \label{tab:taxo}
\end{table}

\subsubsection{An Empirical Ability Taxonomy of LLM}

Table~\ref{tab:taxo} demonstrates the dependency taxonomy between different categories of instructions. 
By performing causal interventions on the distribution of the instruction set, under strict statistical significance criteria, a significant number of dependencies between different categories of instructions could still be identified. For clarity, we define the roots of the taxonomy as \emph{preliminary categories}, the leaf of the taxonomy as \emph{subsequantial categories}, and the intermediate nodes of the taxonomy as \emph{intermediary categories}. In general, the roots of the taxonomy are math and coding-related abilities, such as Python Programming and Math Modeling. Intuitively, these categories correspond to basic reasoning abilities fundamental to completing more complicated tasks. In contrast, categories such the Creativity and Design, Commonsense Understanding, and Communication and Social Media, etc. require multiple capabilities. For example, the creativity generation task requires both abundant knowledge and strong textual generation ability to output creative texts. As a result, these instruction categories serve as ``leaves'' of the taxonomy tree depending on different fundamental abilities. 
The complex dependency patterns indicate that different categories of instructions may not contribute to model performance independently and identically, and suggest the necessity of training LLMs by arranging different categories of instructions in a sequential manner, 
as heuristically, complex skills could be acquired only if the necessary foundation knowledge or ability is equipped.

\section{Category Relationship Guided Instruction Set Optimization}

With the relationship patterns, we further explore optimizing the SFT process of LLM. We investigate optimizing the category distribution of the instruction set based on the correlation pattern between instructions, and concerning the dependency taxonomy between different categories of instructions, we explore optimizing the instruct tuning process by curriculum learning. 

\subsection{Effect Equivalence-based Category Proportion Optimization}

With the correlation patterns between different categories of instructions, we aim to optimize the instruction set by adjusting the proportion of each instruction category. 
The objective function of the optimization could be formalized as:
\begin{equation}
    Obj=s(f_A(w|\{\gamma_{ij}\}_{i,j\in[1,N]}, \mathcal{D}, \mathcal{D}_{\text{candidate}}))
\end{equation}
where $w$ is the optimized weight of each category of instruction, $\gamma_{ij}$ is the equivalence effect coefficient measuring the correlation strength between category $i$ and $j$, $\mathcal{D}$ is the original instruction set, $f_A(\cdot)$ is the weight adjust function, $s(\cdot)$ is a score function evaluating the effectiveness of the weight adjustment. By adjusting the proportion of each category of instructions according to $w$, certain categories of instructions are removed from $\mathcal{D}$, or incorporated into the new instruction set from $\mathcal{D}_{\text{candidate}}$.
However, it could be a challenging task, as the score function is not clearly defined. Generally, it should be related to the performance of LLM finetuned using the adjusted instruction set. Whereas it is rather hard to be modeled using a parametric function and thus obstructs solving $w$. 

To address this issue, we propose an Effect Equivalence-based Category Proportion Optimization (EE-CPO) method. We notice that, since different categories of instructions are correlated, the \emph{equivalent total amount} of instruction category $i$ is not its size $|\mathcal{C}_i|$ alone, but should also include the effect caused by correlation with other categories of instructions:  
$|\mathcal{C}_i|+ \sum_j \gamma_{ji} |\mathcal{C}_j|$.
Note that, $\gamma_{ji}$ could be smaller than 0. Hence, if we can increase the equivalent total amount for an arbitrary category of instructions by adjusting the proportion of instructions meanwhile controlling the total amount of instructions unchanged, then the instruction could be optimized. Thus, the objective of optimization could be formalized as:   
\begin{equation}
    \text{obj}=\max \sum_i |\mathcal{C}_i|+ \sum_j \gamma_{ji} |\mathcal{C}_j| \\ \quad
    \text{s.t.} \sum_i |\mathcal{C}_i| = |\mathcal{D}|    
\end{equation}
where $ |\mathcal{D}|$ is the size of the instruction set $\mathcal{D}$.

Since $\sum_i |\mathcal{C}_i| = |\mathcal{D}|$, by setting $w_j=|\mathcal{C}_j|/|\mathcal{D}|$, i.e., the proportion of category $j$, the objective function could be converted to: 
\begin{equation}
    \text{obj}=\max \sum_i \gamma_{ji} w_j \\\quad 
    \text{s.t.} \sum_j w_j = 1, w_j > 0
\end{equation}

So with this objective function, we can optimize the category proportion. However, this objective function implicitly assumes that all instruction categories have an equal importance. In practice, certain categories would be more important. Hence, another vital issue is how to define the category importance $\alpha_i$. We notice that since the instruction set obtained using previous quality score-based methods achieves promising performance on benchmarks, its category proportion could provide an empirical guide for the category importance. Hence, we estimate $\alpha_i$ using the proportion of category $i$ in the instruction set $\mathcal{D}_{\text{qs}}$ obtained by the quality score-based method, such as DEITA \citep{liumakes}: $\alpha_i=|\mathcal{C}_{\text{qs},i}|/|\mathcal{D}_{\text{qs}}|$.
Thus, concerning the category importance, as shown inFigure~\ref{fig:main_frame}~(d), the objective function could be further formalized as:  
\begin{equation}
    \text{obj}=\max  \sum_i \alpha_j \gamma_{ji} w_j \\ \quad 
    \text{s.t.} \sum_j w_j = 1, w_j > 0
\end{equation}

This objective function could be solved using Linear Programming.  Essentially, the increase of the equivalent total amount could be regarded as increasing the information density of the instruction set. Different from the previous works approaching this by selecting high-quality instructions, EE-CPO achieves this goal by exploiting the correlations between instructions. 




\subsubsection{Experimental Settings}

We constructed three instruction sets, containing 10,000, 20,000, and 50,000 instructions respectively. Given the size of the instruction set and the weight of each category of instruction, the number of each category of instruction could be obtained. For each category, we select the instructions with the highest quality scores from the whole instruction collection. 
The quality score is calculated using the method of \citep{liumakes}. 
To test the generality of our approach, we employ the correlation patterns induced from Qwen-1.5-7B to optimize the instruction set, and test whether the optimized instruction could boost the performance of both the LLama3-8B-base model and Qwen-1.5-7B-base model. Then widely adopted benchmark MT-Bench \citep{zheng2024judging} and AlpacaEval 2.0 \citep{dubois2024length} are used to evaluate the performance of the instruct-tuned LLMs. 

\subsubsection{Baseline Methods}

We make comparisons with the quality score based instruction selection methods: (1) Random Selection selects instances from the whole instruction collection randomly; 
(2) Instag \citep{lu2023instag} measures the informativeness of an instruction instance using the number of Tags it carries; (3) IFD \citep{li2024quantity} evaluates the complexity of instruction using the response loss; (4) DEITA \citep{liumakes} scores the instructions using both a complexity scoring model and a quality score model, then rank the instructions using the synthesis of the quality score and complexity score to select instructions.

\begin{table}
    \centering
    \small
    \begin{tabular}{c|cc|cc}
    \toprule
         & \multicolumn{2}{|c|}{MT-Bench} & \multicolumn{2}{|c}{AlpacaEval2.0} \\
    \midrule
    Method & Qwen1.5 & Llama3 & Qwen1.5 & Llama3 \\
    \midrule
    Random Selection & 6.83 & 6.50 & 5.42 & 5.89 \\
    Instag & 6.69 & 6.96 & 8.90 & 6.67\\
    IFD & 6.53 & 6.25 & 5.77 & 5.43\\
    DEITA (10k) & 7.08 & 7.00 & 8.98 & 7.84\\
    DEITA (50k) & 7.02 & 7.13 & 10.41 & 9.74\\
    \midrule
     EE-CPO (10k) & \textbf{7.09} & \textbf{7.17} & \textbf{9.94} & \textbf{8.09}  \\
     EE-CPO (50k) & \textbf{7.17} & \textbf{7.51} & \textbf{11.29} & \textbf{11.47}  \\
    \bottomrule
    \end{tabular}
    \caption{Performance of Llama3-8B and Qwen 1.5-7B fine-tuned on instruction set obtained by EE-CPO and quality score-based methods.}
    \label{tab:eecpo}
\end{table}

\begin{figure}
    \centering
    \includegraphics[width=1\linewidth]{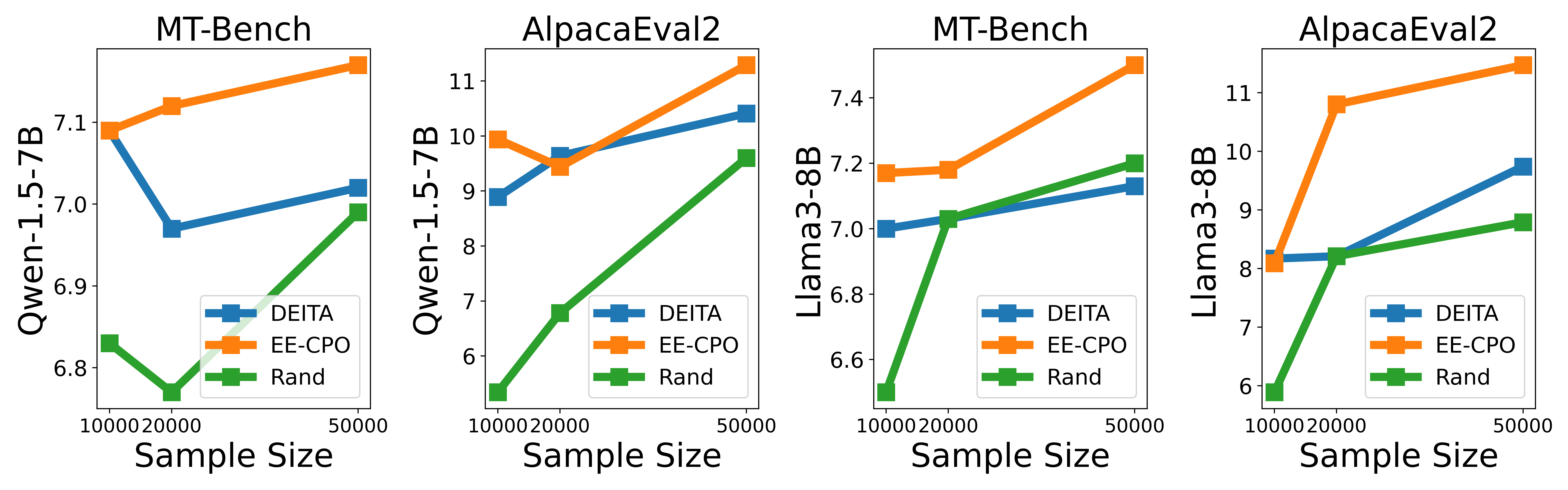}
    \caption{Performance of Llama3-8B and Qwen 1.5-7B fine-tuned on instruction set obtained by EE-CPO and DEITA with different sample sizes.}
    \label{fig:eecpo}
\end{figure}

\subsubsection{Results}
From Table~\ref{tab:eecpo} and Figure~\ref{fig:eecpo}, we observe that:


(1) DEITA outperforms other instruction set optimization methods which selects individual instructions with the highest quality. Compared to the State-of-the-Art method DEITA, our approach EE-CPO could further increase the performance of LLMs by \textbf{only optimizing the proportion of instruction category, without incorporating additional instances}. This shows the necessity of considering the interaction between instructions when optimizing the instruction set and the effectiveness of our approach. Moreover, the advantage of DEITA over random selection diminishes along with the increase of sample size \cite{liumakes}. This is because the number of high-quality instruction is limited. However, our approach demonstrates a consistent advantage over random selection and DEITA on different sample sizes, especially on the instruction set with a relatively large size of 50,000, showing that the necessity of optimizing the distribution of instruction categories could not be offset by enlarging the scale of instruction sets. 

(2) Based on the performance relationship patterns induced on Qwen 1.5, we can improve the performance of both LLama-3 and Qwen 1.5, suggesting the widespread of correlation patterns and the generality of our approach.

(3) Figure 7 of the Appendix shows the weight change of instruction categories. Categories that could not be well substituted by other categories, such as Text Summarization and Academic Writing, together with the preliminary categories, such as Mathematical Modeling and Python Programming, are up-weighted. On the contrary, the instruction categories that can be approximated by other categories of instructions are down-weighted. This suggests the reasonability of our category proportion optimization method.



\subsection{Ability Dependency Taxonomy Guided Curriculum Instruction Learning}

The dependency between instruction categories underscores the need to optimize the SFT process, as learning efficiency would be hindered by the lack of preliminary skills.
To address this issue, we resort to Curriculum Learning. Rather than simply repeating the instruction set with several epochs, Curriculum Learning aims at arranging the samples with different content and difficulty in a sequential manner, so that the model can acquire enough preliminary skills before learning the more complex instructions. 

Specifically, as shown in Figure~\ref{fig:main_frame}~(e), given the dependency taxonomy, and an already existing instruction set $\mathcal{D}$ to equip LLM with enough preliminary skills, we adjust the learning sequence of the SFT process by increasing the proportion of preliminary categories in the early stage of the SFT process. Correspondingly, the proportion of subsequential categories is accordingly decreased. In contrast, at the later stage of SFT, the weight of subsequential categories is increased, and the weight of preliminary categories is decreased, so that the LLM is trained to complete the complex tasks using preliminary skills. For brevity, we abbreviate our proposed approach as DT-CSFT (Dependency Taxonomy guided Curriculum SFT). Thus, DF-CSFT makes adjustments only by adjusting the learning sequential of different categories.


\subsubsection{Experimental Settings}

We obtain $\mathcal{D}$ with 10,000, 20,000, and 50,000 instances using DEITA \citep{liumakes}. As a baseline method, we finetune the LLM on $\mathcal{D}$ with 3 epochs, in other words, the LLM is trained with a total $3|\mathcal{D}|$ instances, with each instance repeated 3 times. In comparison, in the first $|\mathcal{D}|$ instances, DT-CSFT increases the proportion of preliminary category instructions by 50\%. Accordingly, in the last $|\mathcal{D}|$ instances, the proportion of preliminary category instructions is decreased by 50\% by removing them to the first $|\mathcal{D}|$ instance. We also include a baseline (called Mix+) by uniformly mixing additional preliminary category instructions within each epoch. For a dataset with $|\mathcal{D}|$ instances, $2|\mathcal{D}|$ more additional preliminary category instructions are randomly sampled from the whole instruction collection. More details are provided in the Appendix.





\begin{figure}
    \centering
    \includegraphics[width=1\linewidth]{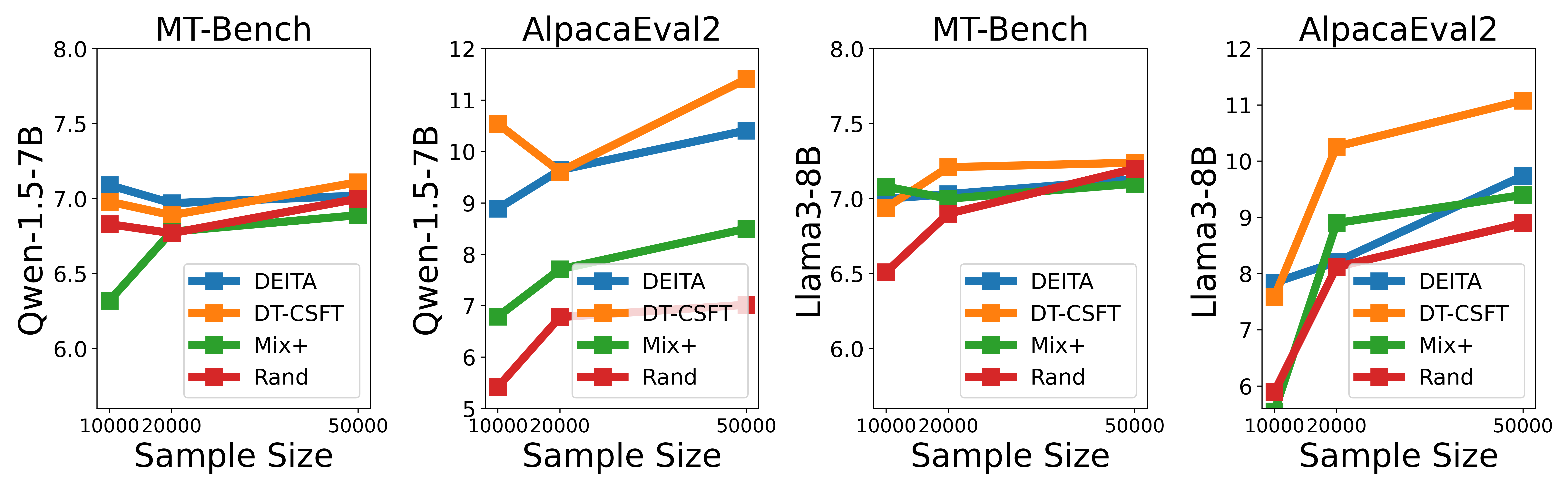}
    \caption{Performance of Llama3-8B and Qwen 1.5-7B fine-tuned on instruction set obtained by DF-CSFT and DEITA with different sample sizes.}
    \label{fig:dfcsft}
\end{figure}

\subsubsection{Results} From Figure~\ref{fig:dfcsft} we observe that:

(1) Compared to the strong baseline DEITA, by \textbf{only adjusting the order of learning different categories of instruction}, DT-CSFT demonstrates improved performance in general. This indicates the reasonability of the taxonomy induced by our approach, as it could provide more necessary fundamental information for LLM to acquire complex skills and thus increase the efficiency of the SFT process. Moreover, based on the taxonomy induced from Qwen 1.5, the performance of LLama-3 could also be improved. This suggests the broad existence and generality of the dependency taxonomy among different LLMs. 

(2) Comparing DT-CSFT with Mix+ shows that incorporating more instructions would not necessarily bring benefits to model performance. This suggests the importance of sequentially arranging the training samples in curriculum learning, as if the preliminary category instructions do not appear in the early stage of SFT, then it could not help to learn the complex skills. 

(3) Our analysis provides theoretical support for the previous empirical observations that Math and Code instructions should be placed in the early stage of SFT \cite{dong2023abilities,hu2024minicpm}. This is because, Math and Code mainly serve as the necessary primary knowledge for more complex tasks. On the contrary, if placed in the later stage of SFT, the learning of complex knowledge would lack of necessary background and thus limit the effectiveness and efficiency.

\section{Related Work}
With the availability of various instruction sets, one crucial issue is how to optimize the existing instruction sets.
To address this issue, most current methods focus on selecting high-quality instructions to obtain a refined instruction set \citep{latif2024fine,lu2023instag,li2024quantity}. While the ``quality'' of instruction could be a comprehensive concept containing multiple aspects. Pioneer works use proxy indicators such as length and perplexity to evaluate the quality of instructions \citep{huang2021generating, wang2024survey}. However, such indicators would not be enough to comprehensively measure the instructions' quality. Another line of work aims at measuring the complexity of instructions, as there is no need to focus too much on learning simple instructions, while overly difficult instructions cannot be learned. Hence, \citet{zhao2023preliminary} propose to measure the complexity of instructions using the number of nodes within a syntax tree, the number of ability tags for completing the instruction \cite{lu2023instag}, or the difficulty of learning the instruction \cite{li2024quantity}. Experimental results suggest improvements in the efficiency of SFT with the instruction set optimized by these methods. Nevertheless, \citet{liumakes} argue that these methods only measure certain aspects of the quality of instructions. They propose DEITA, which simultaneously employs a complexity score, a grammar and factual quality score to choose new instructions. 

However, emerging evidence suggests interactions and dependencies between different categories of instructions \citep{dongabilities,chen2024skill}.  These findings highlight the necessity of optimizing the category proportion of the instruction set and the learning sequence of the SFT process. Whereas a main obstacle is that the interaction and dependent patterns between different instruction categories are largely unknown. To fill this gap, in this paper, we systemically investigate these patterns and explore optimizing the content distribution and SFT schema with regard to them.

\section{Conclusion}

In this paper, we systemically investigate the correlations and dependency taxonomy between different categories of instructions. Analyses results show the widespread of such interactions across multiple categories of instructions and different LLMs, suggesting the necessity of taking the correlation and dependency in the optimization of the content distribution of the instruction set and learning schema of the SFT process. Hence, we further managed to optimize the category proportion and the learning sequence of the instruction set with regard to the correlation and dependency patterns. The improved performance in turn supports the existence of correlation and dependency patterns, together with the reasonability of our investigation and instruction set optimization method. Considering the numerous categories of instruction data, their interaction patterns could be quite complex.
Our work might serve as a pioneer and call for further research to conduct a more comprehensive exploration.

\bibliographystyle{aaai}
\bibliography{aaai}

\begin{thebibliography}{}

\bibitem[\protect\citeauthoryear{Achiam \bgroup et al\mbox.\egroup }{2023}]{achiam2023gpt}
Achiam, J.; Adler, S.; Agarwal, S.; Ahmad, L.; Akkaya, I.; Aleman, F.~L.; Almeida, D.; Altenschmidt, J.; Altman, S.; Anadkat, S.; et~al.
\newblock 2023.
\newblock Gpt-4 technical report.
\newblock {\em arXiv preprint arXiv:2303.08774}.

\bibitem[\protect\citeauthoryear{Chen \bgroup et al\mbox.\egroup }{2024}]{chen2024skill}
Chen, M.; Roberts, N.; Bhatia, K.; Wang, J.; Zhang, C.; Sala, F.; and R{\'e}, C.
\newblock 2024.
\newblock Skill-it! a data-driven skills framework for understanding and training language models.
\newblock {\em Advances in Neural Information Processing Systems} 36.

\bibitem[\protect\citeauthoryear{Dong \bgroup et al\mbox.\egroup }{}]{dongabilities}
Dong, G.; Yuan, H.; Lu, K.; Li, C.; Xue, M.; Liu, D.; Wang, W.; Yuan, Z.; Zhou, C.; and Zhou, J.
\newblock How abilities in large language models are affected by supervised fine-tuning data composition.

\bibitem[\protect\citeauthoryear{Dong \bgroup et al\mbox.\egroup }{2023}]{dong2023abilities}
Dong, G.; Yuan, H.; Lu, K.; Li, C.; Xue, M.; Liu, D.; Wang, W.; Yuan, Z.; Zhou, C.; and Zhou, J.
\newblock 2023.
\newblock How abilities in large language models are affected by supervised fine-tuning data composition.
\newblock {\em arXiv e-prints}  arXiv--2310.

\bibitem[\protect\citeauthoryear{Dubey \bgroup et al\mbox.\egroup }{2024}]{dubey2024llama}
Dubey, A.; Jauhri, A.; Pandey, A.; Kadian, A.; Al-Dahle, A.; Letman, A.; Mathur, A.; Schelten, A.; Yang, A.; Fan, A.; et~al.
\newblock 2024.
\newblock The llama 3 herd of models.
\newblock {\em arXiv e-prints}  arXiv--2407.

\bibitem[\protect\citeauthoryear{Dubois \bgroup et al\mbox.\egroup }{2024}]{dubois2024length}
Dubois, Y.; Galambosi, B.; Liang, P.; and Hashimoto, T.~B.
\newblock 2024.
\newblock Length-controlled alpacaeval: A simple way to debias automatic evaluators.
\newblock {\em arXiv e-prints}  arXiv--2404.

\bibitem[\protect\citeauthoryear{Hahsler, Piekenbrock, and Doran}{2019}]{hahsler2019dbscan}
Hahsler, M.; Piekenbrock, M.; and Doran, D.
\newblock 2019.
\newblock dbscan: Fast density-based clustering with r.
\newblock {\em Journal of Statistical Software} 91(1).

\bibitem[\protect\citeauthoryear{Hu \bgroup et al\mbox.\egroup }{2024}]{hu2024minicpm}
Hu, S.; Tu, Y.; Han, X.; He, C.; Cui, G.; Long, X.; Zheng, Z.; Fang, Y.; Huang, Y.; Zhao, W.; et~al.
\newblock 2024.
\newblock Minicpm: Unveiling the potential of small language models with scalable training strategies.
\newblock {\em arXiv preprint arXiv:2404.06395}.

\bibitem[\protect\citeauthoryear{Huang and Chang}{2021}]{huang2021generating}
Huang, K.-H., and Chang, K.-W.
\newblock 2021.
\newblock Generating syntactically controlled paraphrases without using annotated parallel pairs.
\newblock In {\em Proceedings of the 16th Conference of the European Chapter of the Association for Computational Linguistics: Main Volume},  1022--1033.

\bibitem[\protect\citeauthoryear{Huang and Chang}{2023}]{huang2023towards}
Huang, J., and Chang, K. C.-C.
\newblock 2023.
\newblock Towards reasoning in large language models: A survey.
\newblock In {\em Findings of the Association for Computational Linguistics: ACL 2023},  1049--1065.

\bibitem[\protect\citeauthoryear{Latif and Zhai}{2024}]{latif2024fine}
Latif, E., and Zhai, X.
\newblock 2024.
\newblock Fine-tuning chatgpt for automatic scoring.
\newblock {\em Computers and Education: Artificial Intelligence} 6:100210.

\bibitem[\protect\citeauthoryear{Li \bgroup et al\mbox.\egroup }{2024}]{li2024quantity}
Li, M.; Zhang, Y.; Li, Z.; Chen, J.; Chen, L.; Cheng, N.; Wang, J.; Zhou, T.; and Xiao, J.
\newblock 2024.
\newblock From quantity to quality: Boosting llm performance with self-guided data selection for instruction tuning.
\newblock In {\em Proceedings of the 2024 Conference of the North American Chapter of the Association for Computational Linguistics: Human Language Technologies (Volume 1: Long Papers)},  7595--7628.

\bibitem[\protect\citeauthoryear{Liu \bgroup et al\mbox.\egroup }{2023}]{liumakes}
Liu, W.; Zeng, W.; He, K.; Jiang, Y.; and He, J.
\newblock 2023.
\newblock What makes good data for alignment? a comprehensive study of automatic data selection in instruction tuning.
\newblock In {\em The Twelfth International Conference on Learning Representations}.

\bibitem[\protect\citeauthoryear{Longpre \bgroup et al\mbox.\egroup }{2023}]{longpre2023flan}
Longpre, S.; Hou, L.; Vu, T.; Webson, A.; Chung, H.~W.; Tay, Y.; Zhou, D.; Le, Q.~V.; Zoph, B.; Wei, J.; et~al.
\newblock 2023.
\newblock The flan collection: Designing data and methods for effective instruction tuning.
\newblock In {\em International Conference on Machine Learning},  22631--22648.
\newblock PMLR.

\bibitem[\protect\citeauthoryear{Lu \bgroup et al\mbox.\egroup }{}]{lu2023instag}
Lu, K.; Yuan, H.; Yuan, Z.; Lin, R.; Lin, J.; Tan, C.; Zhou, C.; and Zhou, J.
\newblock \# instag: Instruction tagging for analyzing supervised fine-tuning of large language models.
\newblock In {\em NeurIPS 2023 Workshop on Instruction Tuning and Instruction Following}.

\bibitem[\protect\citeauthoryear{Ouyang \bgroup et al\mbox.\egroup }{2022}]{ouyang2022training}
Ouyang, L.; Wu, J.; Jiang, X.; Almeida, D.; Wainwright, C.; Mishkin, P.; Zhang, C.; Agarwal, S.; Slama, K.; Ray, A.; et~al.
\newblock 2022.
\newblock Training language models to follow instructions with human feedback.
\newblock {\em Advances in neural information processing systems} 35:27730--27744.

\bibitem[\protect\citeauthoryear{Wang \bgroup et al\mbox.\egroup }{2023}]{wang2023self}
Wang, Y.; Kordi, Y.; Mishra, S.; Liu, A.; Smith, N.~A.; Khashabi, D.; and Hajishirzi, H.
\newblock 2023.
\newblock Self-instruct: Aligning language models with self-generated instructions.
\newblock In {\em Proceedings of the 61st Annual Meeting of the Association for Computational Linguistics (Volume 1: Long Papers)},  13484--13508.

\bibitem[\protect\citeauthoryear{Wang \bgroup et al\mbox.\egroup }{2024}]{wang2024survey}
Wang, J.; Zhang, B.; Du, Q.; Zhang, J.; and Chu, D.
\newblock 2024.
\newblock A survey on data selection for llm instruction tuning.
\newblock {\em arXiv e-prints}  arXiv--2402.

\bibitem[\protect\citeauthoryear{Wang, Chen, and Zhu}{2021}]{wang2021survey}
Wang, X.; Chen, Y.; and Zhu, W.
\newblock 2021.
\newblock A survey on curriculum learning.
\newblock {\em IEEE transactions on pattern analysis and machine intelligence} 44(9):4555--4576.

\bibitem[\protect\citeauthoryear{Xiao \bgroup et al\mbox.\egroup }{2023}]{xiao2023c}
Xiao, S.; Liu, Z.; Zhang, P.; and Muennighof, N.
\newblock 2023.
\newblock C-pack: Packaged resources to advance general chinese embedding.
\newblock {\em arXiv preprint arXiv:2309.07597}.

\bibitem[\protect\citeauthoryear{Xu \bgroup et al\mbox.\egroup }{2023}]{xu2023wizardlm}
Xu, C.; Sun, Q.; Zheng, K.; Geng, X.; Zhao, P.; Feng, J.; Tao, C.; and Jiang, D.
\newblock 2023.
\newblock Wizardlm: Empowering large language models to follow complex instructions.
\newblock {\em arXiv e-prints}  arXiv--2304.

\bibitem[\protect\citeauthoryear{Yang \bgroup et al\mbox.\egroup }{2024}]{yang2024qwen2}
Yang, A.; Yang, B.; Hui, B.; Zheng, B.; Yu, B.; Zhou, C.; Li, C.; Li, C.; Liu, D.; Huang, F.; et~al.
\newblock 2024.
\newblock Qwen2 technical report.
\newblock {\em arXiv e-prints}  arXiv--2407.

\bibitem[\protect\citeauthoryear{Yuan \bgroup et al\mbox.\egroup }{2023}]{yuan2023hype}
Yuan, H.; Yuan, Z.; Tan, C.; Huang, F.; and Huang, S.
\newblock 2023.
\newblock Hype: Better pre-trained language model fine-tuning with hidden representation perturbation.
\newblock In {\em The 61st Annual Meeting Of The Association For Computational Linguistics}.

\bibitem[\protect\citeauthoryear{Zhao \bgroup et al\mbox.\egroup }{2023a}]{zhao2023survey}
Zhao, W.~X.; Zhou, K.; Li, J.; Tang, T.; Wang, X.; Hou, Y.; Min, Y.; Zhang, B.; Zhang, J.; Dong, Z.; et~al.
\newblock 2023a.
\newblock A survey of large language models.
\newblock {\em arXiv e-prints}  arXiv--2303.

\bibitem[\protect\citeauthoryear{Zhao \bgroup et al\mbox.\egroup }{2023b}]{zhao2023preliminary}
Zhao, Y.; Yu, B.; Hui, B.; Yu, H.; Huang, F.; Li, Y.; and Zhang, N.~L.
\newblock 2023b.
\newblock A preliminary study of the intrinsic relationship between complexity and alignment.
\newblock {\em arXiv e-prints}  arXiv--2308.

\bibitem[\protect\citeauthoryear{Zheng \bgroup et al\mbox.\egroup }{2024}]{zheng2024judging}
Zheng, L.; Chiang, W.-L.; Sheng, Y.; Zhuang, S.; Wu, Z.; Zhuang, Y.; Lin, Z.; Li, Z.; Li, D.; Xing, E.; et~al.
\newblock 2024.
\newblock Judging llm-as-a-judge with mt-bench and chatbot arena.
\newblock {\em Advances in Neural Information Processing Systems} 36.

\end{thebibliography}

\section{Reproducibility Checklist}
\newcommand{\answerYes}[1][]{\textcolor{blue}{[Yes] #1}}
\newcommand{\answerNo}[1][]{\textcolor{orange}{[No] #1}}
\newcommand{\answerNA}[1][]{\textcolor{gray}{[N/A] #1}}

This paper:

\begin{itemize}
    \item Includes a conceptual outline and/or pseudocode description of AI methods introduced. \answerYes{, see section Introduction}.
    \item Clearly delineates statements that are opinions, hypotheses, and speculations from objective facts and results. \answerYes{, see section Introduction.}
    \item Provides well-marked pedagogical references for less-familiar readers to gain the background necessary to replicate the paper. \answerYes{}
\end{itemize}
\vspace{3mm}

Does this paper make theoretical contributions? \answerYes{}
\vspace{3mm}

Does this paper rely on one or more datasets? \answerYes

\vspace{3mm}

Does this paper include computational experiments? \answerYes{}

\begin{itemize}
    \item  Any code required for pre-processing data is included in the appendix. \answerYes{}
    \item All source code required for conducting and analyzing the experiments is included in a code appendix. \answerYes{}
     \item All source code required for conducting and analyzing the experiments will be made publicly available upon publication of the paper with a license that allows free usage for research purposes. \answerYes{}
    \item All source code implementing new methods have comments detailing the implementation, with references to the paper where each step comes from \answerYes{}
    \item If an algorithm depends on randomness, then the method used for setting seeds is described in a way sufficient to allow replication of results. \answerYes{}
    
    \item This paper specifies the computing infrastructure used for running experiments (hardware and software), including GPU/CPU models; amount of memory; operating system; names and versions of relevant software libraries and frameworks. \answerYes{}
    \item This paper formally describes evaluation metrics used and explains the motivation for choosing these metrics. \answerYes{}
    \item This paper states the number of algorithm runs used to compute each reported result. \answerYes{}
    \item Analysis of experiments goes beyond single-dimensional summaries of performance (e.g., average; median) to include measures of variation, confidence, or other distributional information. \answerYes{}
    \item The significance of any improvement or decrease in performance is judged using appropriate statistical tests (e.g., Wilcoxon signed-rank). \answerYes{}
    \item This paper lists all final (hyper-)parameters used for each model/algorithm in the paper’s experiments. \answerYes{}
    \item This paper states the number and range of values tried per (hyper-) parameter during development of the paper, along with the criterion used for selecting the final parameter setting. \answerYes{}
\end{itemize}

\clearpage
\section{Appendix}

\subsection{Comparison between Correlation Pattern Induced from Different LLMs}

\begin{figure}
    \centering
    \includegraphics[width=0.95\linewidth]{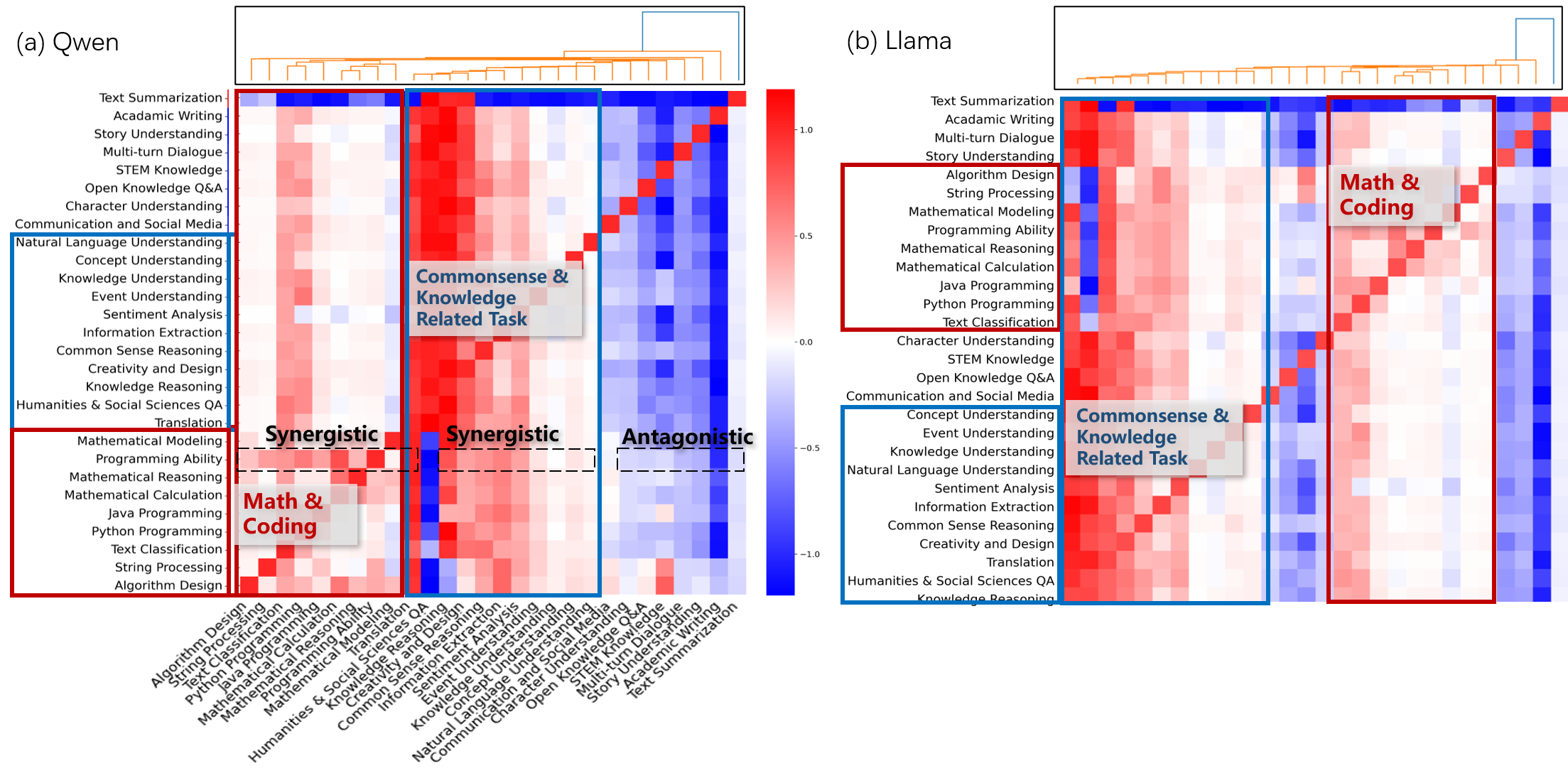}
    \caption{The effect equivalence coefficients between different categories of instructions derived by Qwen and Llama, respectively.}
    \label{fig:substitute_supp}
\end{figure}
Figure~\ref{fig:substitute_supp} shows the correlation pattern derived from Qwen-1.5-7B and Llama3-8B, respectively. Correlation patterns derived from different LLMs demonstrate a high similarity. This shows the widespread and generality of the correlation patterns. Such similarity would partly be brought by the significant overlap between the pretraining of different LLMs, and the inherent logical and knowledge relationships between different categories of instructions.

\subsection{Comparison between Ability Dependency Taxonomy Induced from Different LLMs}

\begin{table*}[ht]
    \centering
    \tiny
        \begin{tabular}{p{2cm}|p{6cm}|p{6cm}}
        \toprule
        & Qwen & Llama \\
        \midrule
        Subsequantial categories & \cellcolor{mydarkblue} \textcolor{white}{['Humanities \& Social Sciences QA', 'Commonsense Understanding', 'Open Domain QA', 'Communication \& Social Media',  'Character Understanding and Role-Playing', 'Creative Writing']} & \cellcolor{mydarkblue} \textcolor{white}{['Humanities \& Social Sciences QA', 'Communication \& Social Media',  'Character Understanding and Role-Playing', 'Creative Writing']} \\
        \midrule
        Intermediary Categories & \cellcolor{mylightgreen} ['Data Process and Analysis', 'STEM Knowledge QA', 'Commonsense Reasoning', 'Concept Understanding', 'Logical Reasoning', 'Information Extraction', 'Sentiment Analysis', 'Story Understanding', 'Text Classification', 'NLU', 'Text Summarization', 'Translation', 'Event Understanding', 'Multiturn Dialogue', 'String Process', 'Academic Writing'] & \cellcolor{mylightgreen} ['Data Process and Analysis', 'STEM Knowledge QA', 'Commonsense Reasoning', 'Concept Understanding', 'Information Extraction', 'Sentiment Analysis', 'Story Understanding', 'Text Classification', 'NLU', 'Text Summarization', 'Translation', 'Event Understanding', 'Multiturn Dialogue', 'String Process', 'Academic Writing', 'Commonsense Understanding', 'Open Domain QA', 'Mathematical Modelling'] \\
        \midrule
        Preliminary Categories & \cellcolor{mylightgray} ['Math Reasoning', 'Mathematical Modelling', 'Arithmetic Calculation', 'Python', 'Java', 'Programm Ability', 'Coding Algorithm'] & \cellcolor{mylightgray} ['Math Reasoning', 'Arithmetic Calculation', 'Python', 'Java', 'Programm Ability', 'Coding Algorithm'] \\
        \bottomrule
        \end{tabular}
    \caption{Dependency taxonomy between instruction categories derived from two LLMs.}
    \label{tab:taxo_supp}
\end{table*}

Table~\ref{tab:taxo_supp} shows the dependency taxonomy derived from Qwen-1.5-7B and Llama3-8B, respectively. Comparison between the two taxonomies suggests a high similarity between two taxonomies, especially in the preliminary categories. This shows the widespread and generality of such dependency taxonomy, as the dependency patterns are essentially decided by the inherent relationship of knowledge and skills for completing different categories of instructions.

\begin{table}
    \centering
    \begin{tabular}{cc}
    \toprule
    Alpaca GPT4     & LIMA  \\
    \midrule
    Alpaca GPT4 ZH     & LongForm  \\
    \midrule
    BaiZe   &  logi-COT \\
    \midrule
    BELLE Generated Chat     & ShareGPT-Chinese-English-90k  \\
    \midrule
    BELLE Multiturn Chat     & UltraChat  \\
    \midrule
    BELLE train 3.5M CN     & Wizard Evol instruct zh  \\
    \midrule
    databricks-dolly-15K     & Wizard Evol instruct 196K \\
    \midrule
    BELLE School Math & Code Alpaca 20K \\
    \midrule
    MetaMath & WildChat \\
    \midrule
    COIG-CQIA  &  \\
    \bottomrule
    \end{tabular}
    \caption{List of instructions included for analysis.}
    \label{tab:dataset}
\end{table}

\subsection{Instruction Collection}

In this paper, before analyzing the interaction relationships between instruction categories, a prerequisite is collecting enough instructions so that the main categories of instructions can be covered. provide a comprehensive list. Based on their list, we exclude all instructions that are not constructed by human annotation or advanced LLMs such as GPT-4 or ChatGPT. Beyond their list, we also include Logi-QA, Wild-Chat, and COIG-CQIA. Table~\ref{tab:dataset} provided a detailed list of the included instruction set.

Given the instruction set, we employ SimHash with a threshold=0.95 to remove the potential duplicated instructions. After the duplication process, 9,509,526 instances in total are left in the instruction collection.

\subsection{Tag Generation}

Considering the vast amount of instruction, we construct an automatic tagging system that employs an LLM to generate tags for a given instruction. 

Specifically, given an instance from the instruction collection which is composed by (several) \{Instruct-Response\} pair(s), we concatenate them into a string, and as Figure~\ref{fig:prompt} shows, using the following prompt, to demand the LLM to generate tags describing the necessary knowledge and skills for completing the dialogue described by the \{Instruct-Response\} pair(s):

\begin{figure*}
    \centering
    \includegraphics[width=0.8\linewidth]{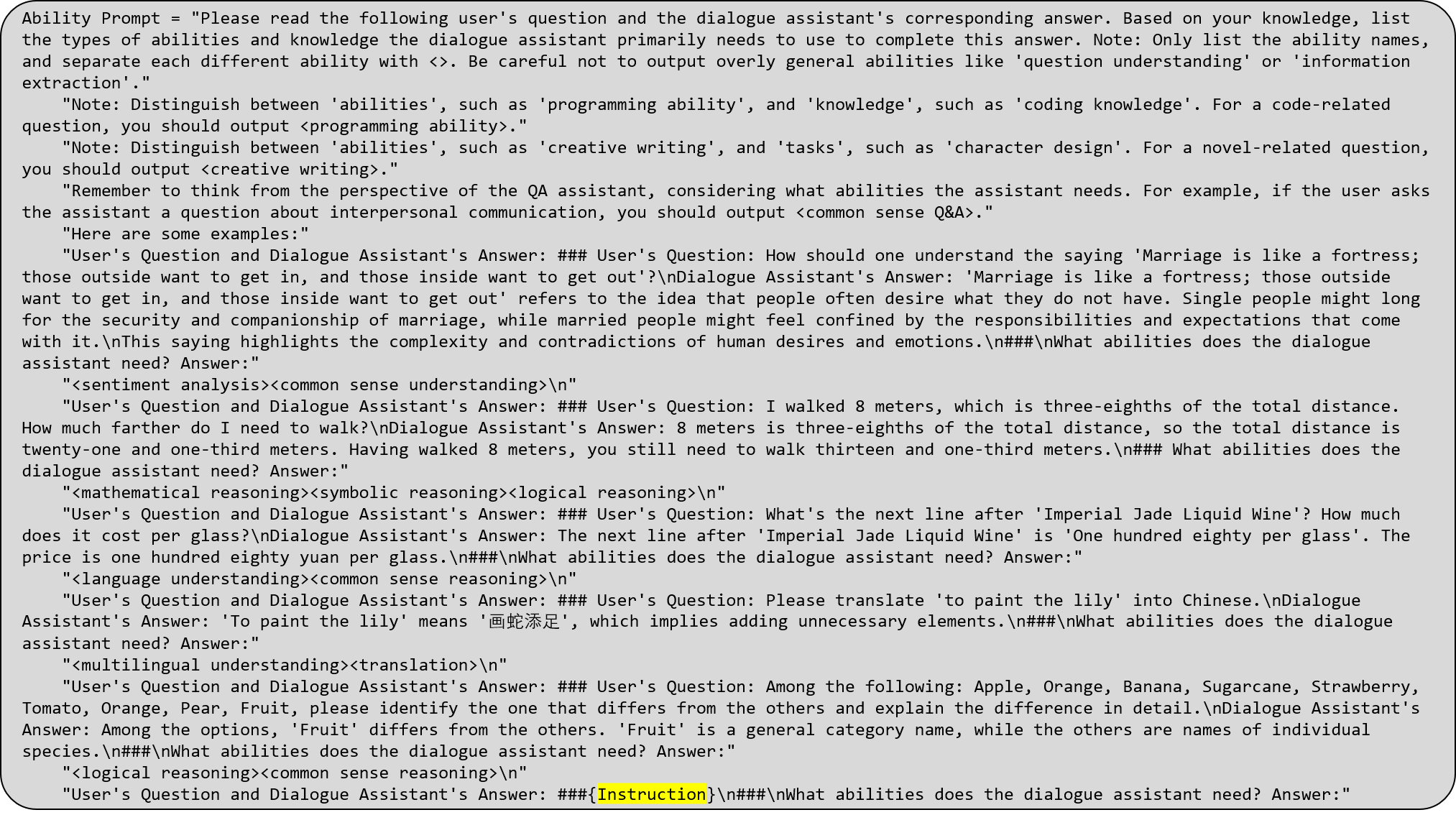}
    \caption{Prompt used for guiding the LLM to generate tags for given instruction.}
    \label{fig:prompt}
\end{figure*}

We employ Qwen-1.5-72B-instruct as the tagger.

\subsection{Tag Normalization}

The LLM would describe one kind of knowledge or skill with different expressions, for example, ``math calculation'' with ``mathematical calculation''. To address this issue, we propose to combine these tags according to their semantic similarity. Specifically, we obtain the embedding of the tags using BGE \citep{xiao2023c}. Then semantically similar tags are recognized if their cosine similarity of embeddings is larger than an empirical threshold $\lambda$ = 0.85. For a set of semantically similar tags, they are normalized to the one with the highest frequency among them. After the normalization process, tags With a frequency lower than 100 are considered as long-tail and are filtered out \cite{lu2023instag}. After the normalization and filtering, a total of 21,378 tags are left. We manually selected 29 categories across 7 domains for analysis. 

\begin{figure*}[h]
    \centering
    \includegraphics[width=0.7\linewidth]{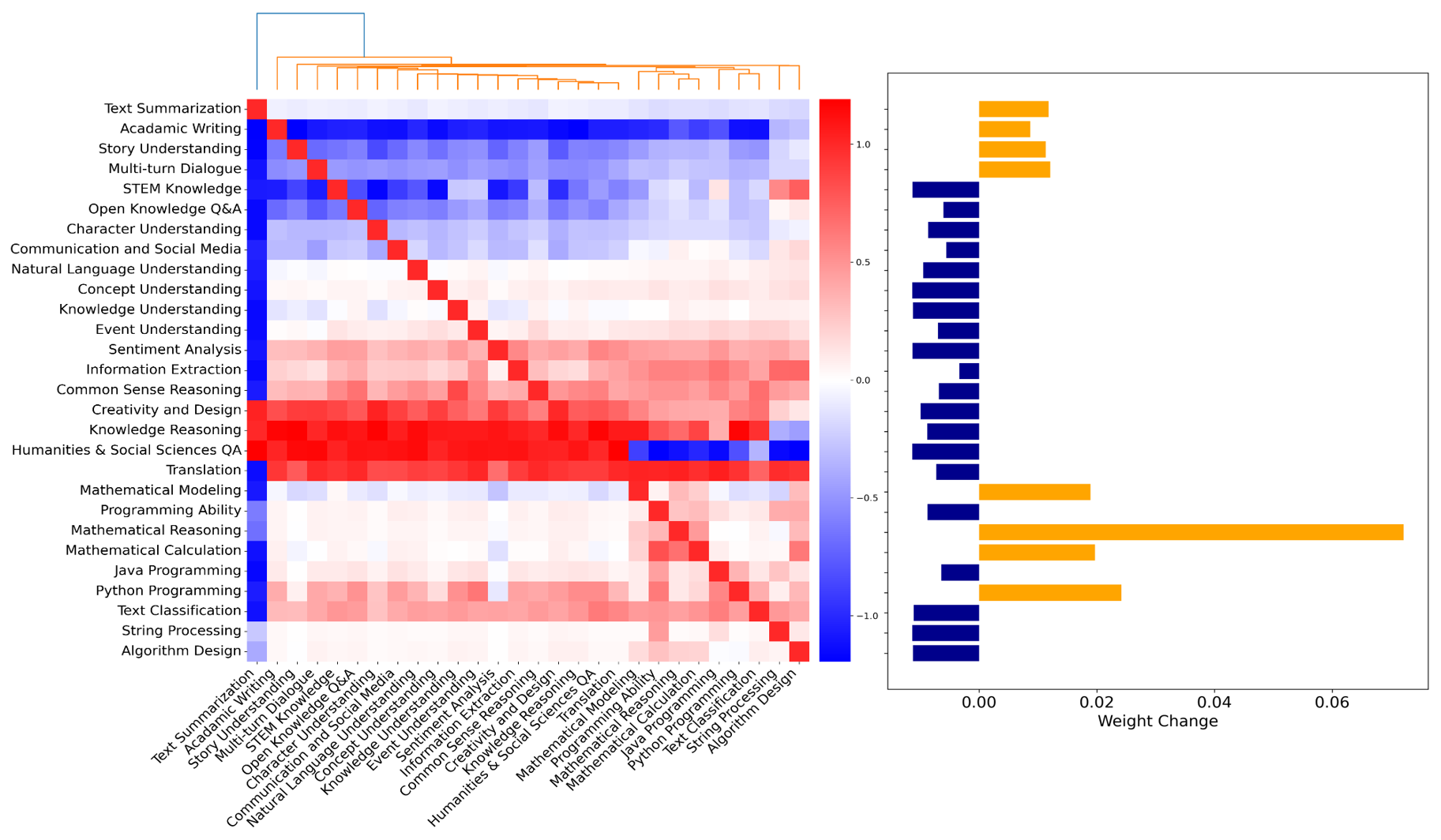}
    \caption{Change in weights of categories after optimization.}
    \label{fig:weight_change}
\end{figure*}

\subsection{Experimental Settings}


Before analysis, we excluded the instructions that were semantically similar to the test instances in the benchmarks AlpacaEval and MTBench using semantic similarity. Instructions with a cosine similarity larger than 0.3 are excluded from further analysis. Without generality, we only select instructions in English for analysis. 

In the causal intervention-based instruction correlation analysis, the base instruction set $\mathcal{D}$ are constructed by randomly sampling 1,000 instances of each category from the whole instruction collection. The same base instruction set is also adopted in the ability taxonomy induction, to control potential confounders. 

To induce the correlation pattern of each category of instruction with the others, at each time, 2,000 instructions of one category are added into the base instruction set to obtain $\mathcal{D}\cup \mathcal{C}_i$. Then we use $\mathcal{D}\cup \mathcal{C}_i$ to fine-tune an LLM $M$ and obtain $M^{\cup \mathcal{C}_i}$. During the fine-tuning process, $M$ is fine-tuned for 3 epochs, with a batch size of 32, the initial learning rate of 9.65e-6,  optimized with the Adam optimizer, $\beta_1=0.9$, $\beta_2=0.95$ \cite{ouyang2022training}. This group of hyperparameters is also adopted in all other sections when the fine-tuning process is involved. 

For the baseline quality score-based instruction selection method IFD and Instag, we implemented their using their codes on our instruction collection. As we included more instructions in the collection set, our implementations outperformed the original implementations on the MT-Bench and AlpacaEval2.0.

In the ability dependency taxonomy guided curriculum instruction learning, assume the base instruction set  $\mathcal{D}$ contains $N_{pre}$, $N_{inter}$, and $N_{sub}$ preliminary, intermediate, and subsequential category instructions. Note that $\mathcal{D}=N_{pre}+N_{inter}+N_{sub}$. Then the curriculum data is arranged as follows:

\begin{itemize}
    \item First $\mathcal{D}$ instances=

random\_shuffle($1.5N_{pre}\text{[preliminary category instructions]}+$

$N_{inter}\text{[intermediary category instructions]}+$

$(N_{sub}-0.5N_{pre})\text{[subsequential category instructions]}$)

\item Second $\mathcal{D}$ instances=

random\_shuffle($N_{pre}\text{[preliminary category instructions]}+$

$N_{inter}\text{[intermediary category instructions]}+$

$(N_{sub}-0.5N_{pre})\text{[subsequential category instructions]}$)

\item Third $\mathcal{D}$ instances=

random\_shuffle($0.5N_{pre}\text{[preliminary category instructions]}+$

$N_{inter}\text{[intermediary category instructions]}+$

$(N_{sub}-1.5N_{pre})\text{[subsequential category instructions]}$)
 
\end{itemize}

In this way, we arranged the instructions to increase the proportion of preliminary category instructions in the early stage of SFT and increase the proportion of subsequential category instructions in the later stage of SFT, meanwhile keeping the total number and constitution of the curriculum data identical to the original base instruction set $\mathcal{D}$ unchanged.

\subsection{Changes in weights of Categories after Optimization}

As Figure~\ref{fig:weight_change} shows, in general, the weights of categories that can be well substituted such as NLU, Concept Understanding, or Commonsense Reasoning are decreased. On the contrary, the weights of Text Summarization, Academic Writing which could not be well substituted by another category of instructions, and the preliminary categories, such as Mathematical Modeling and mathematical Reasoning, are increased.

\end{document}